\documentclass[preprint,12pt,authoryear,preprint, review]{elsarticle}

\usepackage{subcaption}
\usepackage[usenames]{color}
\definecolor{color}{RGB}{255,0,0}
\usepackage{soul}
\usepackage{url}

\journal{Pattern Recognition}

\begin{document}
	
	\begin{frontmatter}
		
		
		
		\title{A Hybrid Approach for Tracking Individual Players in Broadcast Match Videos}
		
		
		\author[label1]{Roberto L. Castro}
		\author[label1]{Diego Andrade}
		\author[label1]{Basilio Fraguela}
		\address[label1]{Universidade da Coru\~na, CITIC, Computer Architecture Group. A Coru\~na, Spain}

		\address{}
		
		\begin{abstract}
			Tracking people in a video sequence is a challenging task that has been approached from many perspectives.
			This task becomes even more complicated when the person to track
			is a player in a broadcasted sport event, the reasons being the existence of difficulties such as frequent camera movements or switches, total and partial occlusions between players, and blurry frames due to the codification algorithm of the video. 
			This paper introduces a player tracking solution which is both fast and accurate. This allows to track a player precisely in real-time. The approach combines several models that are executed concurrently in a relatively modest hardware, and whose accuracy has been validated against hand-labeled broadcast video sequences. Regarding the accuracy, the tests show that the area under curve (AUC) of our approach is around 0.6, which is similar to generic state of the art solutions. As for  performance, our proposal can process high definition videos (1920x1080 px) at 80 fps.
		\end{abstract}
		
		
		
			
			
			
		
	\end{frontmatter}
	\newpage
	
	\section{Introduction}
	\label{sec:intro}
	
	Object tracking is one of the most important problems in computer vision. The objective is to track a specific object in a video sequence. To do this, it can take advantage of the initial position of the object specified by a human operator, and afterwards, of the position of the object in the previous frame. 
	In machine learning (ML), this problem has been usually faced using an approach known as tracking by detection. This method detects the object in the frame without taking into account its previous position. This leads to  
	a high computational cost, since it cannot easily determine a constrained region of interest (ROI) where the tracked object is likely to be located. As a result, it is not feasible to track objects at real-time following this approach. For instance, if the video source is a high definition video of 60 fps, each detection has to process a high-definition frame in just 16 ms. Also, object detection might be successful in only a fraction of the frames of the video sequence, as the tracked 
	object might not detectable in 
	every frame.
	
	This paper presents a solution that attempts to overcome these limitations by combining costly but accurate deep learning (DL) techniques with faster but less accurate math-based algorithms, obtaining a resulting solution which is both accurate and fast at the same time. This solution is applied to the challenging case study of tracking an individual player in a broadcasted sport event. The special difficulties introduced by this case study~(\cite{cite11}) are the existence of frequent camera movements and switches,  partial and total occlusions of the tracked object, players of the same team with a very similar appearance to the tracked player, blurry images due to the codification algorithm, and video intervals where the tracked player is out of camera. 
	
	The proposed solution combines two deep neural networks (DNN), Faster-RCNN and SSD, with the Kernelized Correlation Filter (KCF) algorithm. The three models are executed concurrently, although with different frequencies of execution, exchanging 
	information to build a hybrid tracker. This hybrid approach 
	does not need a human operator or supervisor that makes the selection of the initial position of the tracked player. This initial selection is automatically made by the Faster-RCNN network, which is trained to detect a specific player. The same kind of training is applied to SSD. The datasets required to train both networks were generated using a semiautomated tool that uses OpenCV object tracking algorithm. 
	These DNNs 
	allows the tracker to recover from situations where the player is out of camera, as well as from temporary detection errors.
	
	The results show that this approach is able to track a specific player through complicated and challenging video sequences, containing several of the situations that make this case study specially complicated. The experiments show that the proposed algorithm can be executed more than 60~times\footnote{80 times on average} per second on the frames of a high definition video. The accuracy of the tracker has been validated against hand-labeled video sequences obtaining a value of the area under curve (AUC) of around 0.6.
	
	The rest of this paper is organized as follows.
	Section~\ref{sec:motiv} motivates this paper.
	Section~\ref{sec:ot} formalizes the problem of object tracking and our case study, and introduces some of the solutions available.
	Section~\ref{sec:ourapp} introduces the solution provided in this paper. Then, Section~\ref{sec:exp} presents the experimental results, while Section~\ref{sec:conc} presents our conclusions and future work.

	\section{Motivation}
	\label{sec:motiv}
	
	As mentioned in the introduction, object tracking is a difficult task for which a big number of solutions have been proposed. None of them has proved to be both fast and accurate enough for the case study of this paper, the unsupervised tracking of an individual player in a broadcasted sport event. This is related to the several complications introduced by this case study, which have already been mentioned in Section~\ref{sec:intro}. 
	
	We tried a number of existing solutions to solve this problem before creating our own one. First, we tried several classifiers based on the ``tracking by detection'' approach and implemented in the OpenCV library~(\cite{opencv}). 
	Multiple Instance Learning~(\cite{mil}) (MIL) and Kernelized Correlation Filters (KCF)~(\cite{kcf}) are the two that provided the best results in the case study.
	This kind of methods train a discriminative classifier in an online manner to separate the object from the background, using the current tracker location to extract positive and negative samples from the current frame. When the tracking is not precise, the appearance model uses a sub-optimal positive sample, which can degrade its accuracy 
	over time and cause drift\footnote{In this context, the term is used to indicate the progressive loss of the tracked object due to error accumulation}. Some of these methods take negative and positive samples from the areas of the frame around the location of the tracker to attenuate this problem.  The accuracy of these trackers in our case study is extremely sensitive to the quality of the initial selection, and they can only track the player for a few seconds. After that, these trackers suffer drifting or lose the tracked object in a sudden. 
	
	An alternate solution was to use Convolutional Neural Networks (CNNs) to track the player. These networks are trained offline to track an individual player. For this, the model should be fed with thousands 
	of sample frames indicating the location of the tracked player. Again, these trackers are based on the approach of ``tracking by detection'', as they treat every frame as a new problem where the player has to be located without knowing its previous location. This implies that they lose some context information with respect to traditional tracking approaches, which 
	take advantage of the previous locations of the tracked object. This also implies a higher computational cost, as this alternative does not consider a reduced area of interest where the tracked player is likely to be located.
	
	Among these trackers, FasterRCNN generates the best results. To further evaluate its effectiveness, we made a tracking test with a short video containing several challenging situations such as partial or total occlusions or camera switches, for example. For this test, and all the other tests throughout the paper, we are going to consider the following metrics: The precision of the tracking, for every frame, will be evaluated using the average overlap metric, which calculates the IoU (\textit{Intersection Over Union}) between two bounding boxes (BBs), one being the ground-truth region ($\Lambda^{G}$) and the other being the one detected by the tracker to evaluate ($\Lambda^{T}$). This metric is defined using the following formula:
	
	\[
	\Phi (\Lambda^{G}, \Lambda^{T}) = \{\Phi_{t}\}^{N}_{t=1}, \phi = \frac{R_{t}^{G} \cap R_{t}^{T}}{R_{t}^{G} \cup R_{t}^{T}}
	\]
	
	where:
	\begin{itemize}
		\item $R_{t}^{G}$ is the ground-truth bounding box
		\item $R_{t}^{T}$ is the predicted bounding box
		\item $R_{t}^{G} \cap R_{t}^{T}	$ is the area of overlap between both bounding boxes
		\item $R_{t}^{G} \cup R_{t}^{T}	$ is the area encompassed by both bounding boxes
		\item $\Phi (\Lambda^{G}, \Lambda^{T})$ is the division between the area of overlap by the area of union - \textit{Intersection over Union}
	\end{itemize}
	
	
	For this metric, we are considering that an IoU above $0.5$ is successful. The average overlap is a good metric to measure how good is each individual prediction. In order to show the precision on a sequence of frames, we will be using the \textit{One Pass Evaluation} (OPE) curve. This curve shows the percentage of correct frames, correct meaning that their IoU is above a given threshold. The curve shows this percentage for different thresholds. A summary metric of this curve is the  \textit{Area Under the Curve (AUC)} value, which is calculated by computing the integral of the OPE curve. This metric takes a value between 0 and 1 and the higher it is, the better the result is. 
	
	\begin{figure}
		\centering
		\includegraphics[scale=0.6]{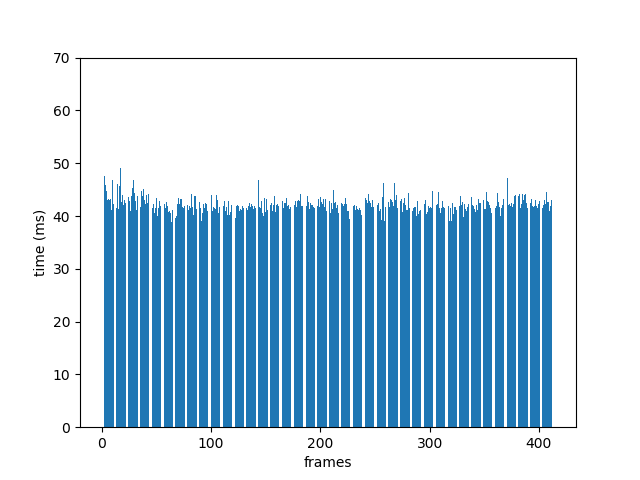}
		\caption{Time required by Faster RCNN to process each frame}\label{fig:fpsfaster}
	\end{figure}
	
	\begin{figure}
		\centering
		\includegraphics[scale=0.6]{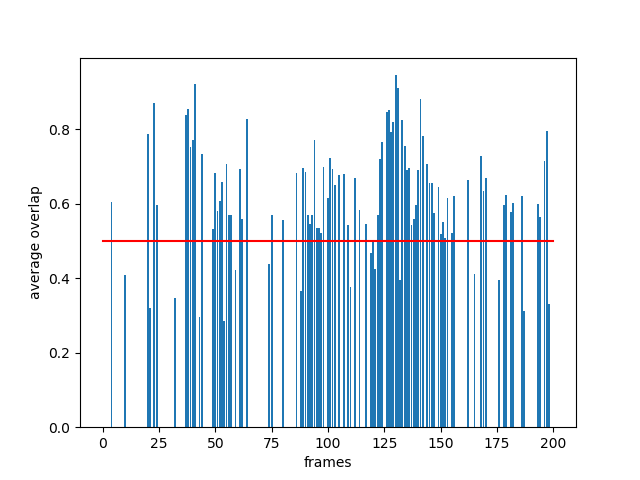}
		\caption{Overlap between hand-labeled frames vs FasterRCNN detection}\label{fig:ovhandvsfaster}
	\end{figure}
	
	\begin{figure}
		\centering
		\includegraphics[scale=0.6]{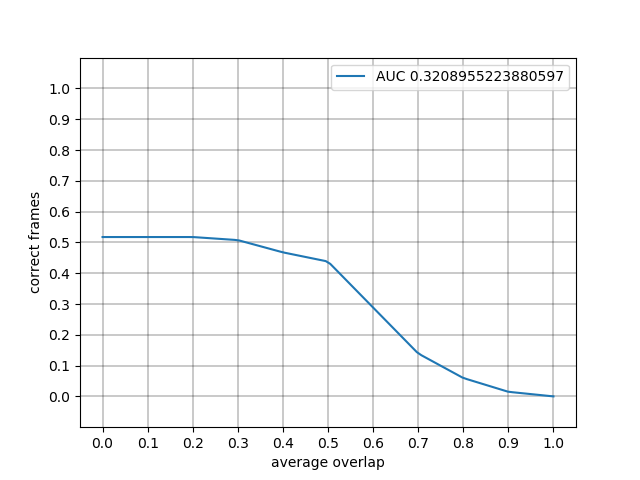}
		\caption{OPE curve of the detection made with FasterRCNN vs hand-labeled frames}\label{fig:opefaster}
	\end{figure}
	
	The video used for this preliminary test has a full HD resolution (1920x1080 px) and a frame rate of 60~frames per second
	(fps), and it has been obtained from the SoccerNet dataset~(\cite{giancola2018soccernet}). The test has been executed on a workstation equipped with an Nvidia GTX 1080 Ti card, an Intel i7 8700k processor and 16 GB of RAM memory. Notice that along this paper, the training and inference processes of deep neural networks will be using all the hardware available in the system, including the GPUs.
	We are interested in measuring the accuracy and the performance of the model.  This way, Figure~\ref{fig:fpsfaster} represents the time required to execute the model in every frame. The execution time is almost constant around 45 ms, which is insufficient for processing our video in 
	real time. For instance, in order to maintain a frame rate of 30 fps we need to process each frame in just 33 ms, while for a frame rate of 60 fps we need to execute it in just 16 ms.

	Figure~\ref{fig:ovhandvsfaster} shows the average overlap between the hand-labeled location of the tracked player and the location detected by Faster-RCNN. As hand-labeling the whole video would have taken too much time, we decided to focus these accuracy results on a subset of 200 frames of the video sequence which were considered specially challenging.
	The outputs of this detector are a bounding box enclosing the most probable location of the tracked object and the confidence of the detection, which is provided with a score between 0 and 1 (the higher the better). This figure assigns an average overlap of 0 to those frames where  this score is below 0.6, as this means that in those cases the confidence of the detector is bellow 60\%. In the figure, a horizontal red line marks the 0.5 threshold for the average overlap, above which the detection is considered successful. This threshold is only surpassed by around 50\% of the video frames, which evidences that the detection provided by FasterRCNN is discrete. 

	Finally, Figure~\ref{fig:opefaster} represents the OPE 
	curve of the detection made with FasterRCNN. This curve shows that this tracker only achieved an overlap (w.r.t. hand-labeled frames) above 50\% in 45\% of the cases, and that there was a significant
	fraction of the frames where no detection at all was made. The Area Under Curve (AUC) value is just 0.32, which is far from the state of the art methods of object detection. 
	
	These results show that techniques based on maths and statistics cannot successfully address our case study. CNNs seem to provide promising results, but their computational cost is excessive and, although they can track the player in some frames, there are 
	many frames where they cannot detect the player at all.

	\section{Object tracking}
	\label{sec:ot}
	
	Object tracking is an important problem in the field of computer/artificial vision. Its aim is to track an object through a sequence of frames of a video. The outcome of a tracking algorithm would be a tracking area around the tracked object. In this work, we have adopted the usual simplification of providing a rectangular bounding box (BB) around the tracked object.

	If our purpose were to track an object on a single frame instead of a video sequence, the problem would be referred as object detection. In some works~(\cite{kcf,mil}), the idea of using an object detection algorithm to track that object across a video sequence has been tried successfully. 
	However, notice that, in the case of object tracking, the history of the previous locations of the object can be used to provide a more accurate and faster tracking. Usually, this information allows establishing a ROI around the immediately previous location of the tracked object, based on the assumption that the movement of the object has a certain locality and that sudden movements from one edge to another of the frame only happen seldom. This ROI can be used to narrow the potential locations of the tracked object.

	Regarding the nature of the target of the tracking, sometimes only an individual unique object needs to be tracked, while in other situations it is required to track all the objects of the same type. Usually, the camera is static and it provides always the same perspective, although sometimes there can be camera switches or movements, which complicates the task.
	
	In this paper, we have focused on tracking an individual player on a broadcasted soccer match. The output for each frame will be a BB around the most probable location of the tracked player. 
	
	Most of the algorithms that can be found in the bibliography have been tested using data sets, which unlike what happens in our case study, track objects that occupy a high percentage of the frame; a situation that allows to extract characteristics that ease the identification of the objects. The OTB benchmarks~(\cite{WuLimYang13}) and the VOT benchmarks~(\cite{VOT_TPAMI}) are examples of this type of data sets. 
	
	Table~\ref{tab:Siam} summarizes the AUC value obtained by recent studies when addressing some of the aforementioned benchmarks.
	The best result corresponds to the the SiamVGG network~(\cite{cite5}), whose AUC value varies between  $0.610$ and $0.654$ for different benchmarks. 
	Regarding performance, the SiamVGG network can process a maximum of 35 fps. These numbers will allow us later to put into perspective the level of accuracy and the performance achieved by our strategy.

	\begin{table}[h]
		\centering
		\footnotesize
		\begin{tabular}{|l|l|l|l|l|}
			\hline
			\textbf{tracker}  & \textbf{OTB-2013} & \textbf{OTB-2015} & \textbf{OTB-100}\\ \hline
			\textbf{SiamFC-3s (\cite{bertinetto2016fully})} & 0.607 & 0.516 & 0.582 \\ \hline
			\textbf{CFNet (\cite{valmadre2017end})} & 0.611 &  0.530 & 0.568\\ \hline
			\textbf{RASNet (\cite{wang2018learning})} & 0.670 &  - & 0.642\\ \hline
			\textbf{SA-Siam  (\cite{he2018twofold})} & 0.677 & 0.610 & 0.657\\ \hline
			\textbf{DSiam (\cite{guo2017learning})} & 0.656 &  - & - \\ \hline
			\textbf{SiamRPN (\cite{li2018high})} & - &  - & 0.637\\ \hline
			\textbf{SiamVGG (\cite{cite5})} & 0.665 &  0.610 & 0.654\\ \hline
		\end{tabular}
		\caption{AUC for some recent studies of object tracking for several popular benchmarks.}
		\label{tab:Siam}
	\end{table}
	
	Player tracking is a sub-field of object tracking where a number of ad-hoc solutions have been proposed. These solutions have approached the problem from different perspectives, and using very different theoretical foundations. The survey~(\cite{cite11}) contains a comprehensive study of all of them. Some of them have been explicitly tested as part of this research work. For instance, we attempted to integrate the idea of using Kalman Filters~(\cite{svensson2010target}) in our hybrid tracker, but it failed to capture camera movements, or situations where players change 
	of direction or speed up suddenly.
	
	Recent implementations of the Meanshift~(\cite{Comaniciu:2002:MSR:513073.513076}) and Camshift~(\cite{Allen:2004:OTU:1082121.1082122}) techniques 
	were also tried but, in the best case, they were only able to track a player for a few seconds in our benchmark video sequences.

	\section{The hybrid player tracker}
	\label{sec:ourapp}
	
	The main challenge of our tracker is to use object detection DNNs to perform efficient and precise object tracking. In order to do this we need: (1) to use the history of previous locations detected by the tracker to define a reduced ROI in each frame on which to make the current detection, and (2) to execute concurrently different algorithms at different frequencies to make a precise tracking of a player in a reasonable time. 
	
\begin{figure}[tb]
		\centering
		\includegraphics[scale=0.3]{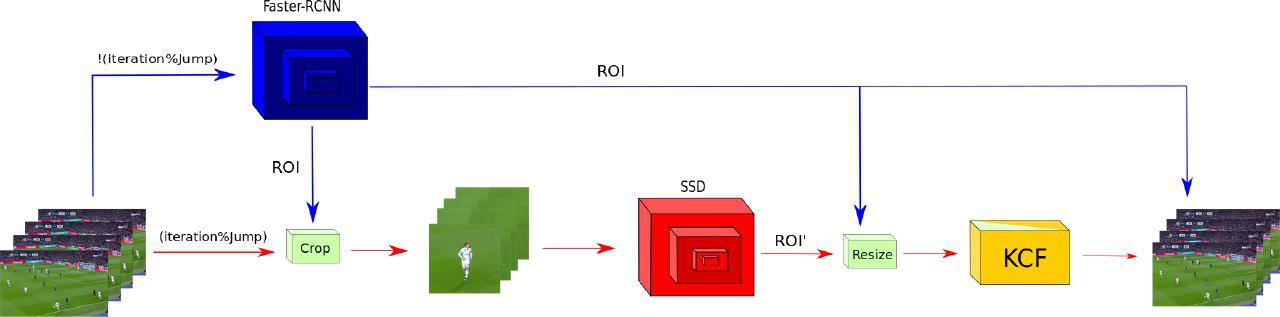}
		\caption{Architecture of the hybrid player tracker}\label{fig:arch}
	\end{figure}

	Figure~\ref{fig:arch} represents the high-level architecture of our solution. In one of every \emph{jump} frames, the object detection Faster-RCNN network is executed to detect the player. The output of this module 
	will be the most likely position of the tracked player at this time. 
	The blue arrows represent the flows of information that happen during the processing of these \emph{jump} frames.
	
	The location given by Faster-RCNN will be used to crop subsequent frames establishing a reduced area of interest where to search for the tracked player. This area of interest will be further refined and updated by a SSD network detector which will be executed for each one of the remaining $jump-1$ frames. Concurrently, 
	the KCF algorithm will be used to do a fine-grain tracking of the object in the ROI on each frame and to provide the output of the algorithm in these frames. The flows of information that happen during the processing of these $jump-1$ frames is represented with red arrows in Figure~\ref{fig:arch}.
	
	\begin{figure}[tb]
		\centering
		\includegraphics[scale=0.35]{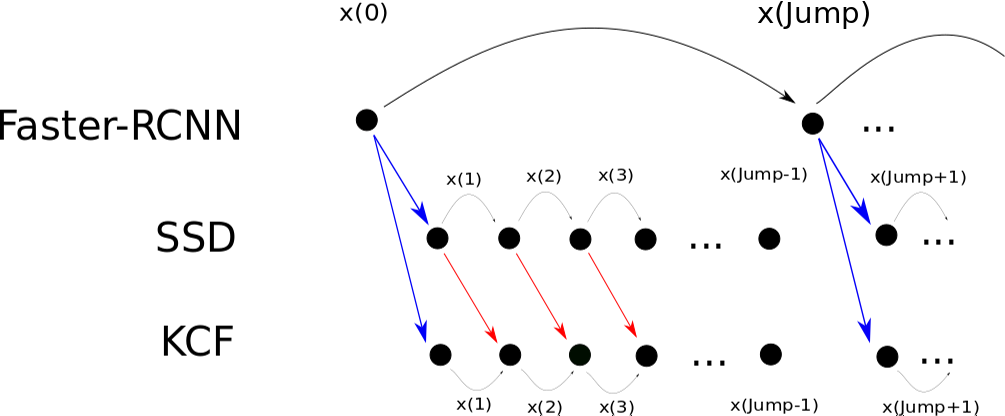}
		\caption{Sequence diagram of the operation of the hybrid player tracker}\label{fig:seq}
	\end{figure}
	
	Figure~\ref{fig:seq} gives a different view of the algorithm representing a timeline of the algorithm execution, where we can appreciate that the three algorithms are executed with different frequencies. In this figure, the color code of Figure~\ref{fig:arch} is maintained. Namely, it also uses blue arrows for the first frame of each group of $jump$ frames and  red arrows for the remaining ones.
	
	The conception of this algorithm resulted from an incremental process driven by a large number of tests. Our initial hypothesis was that executing Faster-RCNN on each frame would be feasible, but as commented in Section~\ref{sec:motiv}, we only obtained a fraction of the required performance to be able to perform the tracking in
	real-time. Also, in that same section we showed that when using this approach, a successful detection was made only in around a 20\% of the frames, which means that the tracking algorithm lacked continuity in the detection. 
	
	Also, the tests made using KCF alone showed that this algorithm requires periodical supervision, as the initial location of the player has to be indicated by a supervisor, and after some time the algorithm suffers drifting, which should be corrected periodically. 
	In our proposal, the role of the supervisor is played by the Faster-RCNN algorithm. However, this supervisor does not intervene only at the beginning of the tracking, but rather each $jump$ frames. The supervision of the SSD detector at every frame is effective avoiding the KCF drifting problem. This is done by adjusting the KCF BB if there is a serious discrepancy between it and SSD. However, adjusting KCF when a discrepancy with SSD is detected for the first time proved to turn the overall solution unstable. Therefore, instead of doing this, we found that the solution works better if the adjustment is only made when both techniques diverge in more than 3 frames in a row. This 
	value was obtained through experimentation.
	
	
	\section{Experimental results}
	\label{sec:exp}
	
	This section includes: (1) the accuracy and performance results of our approach, and (2), the data of the traning required for the deep neural networks used in our solution. They are discussed in~\ref{sec:acc} and~\ref{sec:training} respectively.
	
	\subsection{Accuracy and performance}\label{sec:acc}
	The accuracy and performance of our hybrid tracker was tested by means of experiments whose purpose was to track a total of 4~players from 2~different soccer teams. For each player we chose a specially challenging video sequence containing all the situations that may complicate the tracking, and that were commented in Section~\ref{sec:intro}. There are small variations in the duration of each video sequence selected for each player. The reason is that we tried to balance the videos so that in all of them the tracked player is on camera a similar amount of time.
	The videos used in these experiments have been obtained from the SoccerNet dataset (\cite{giancola2018soccernet}). 
	

	\begin{figure}
		\begin{subfigure}{\linewidth}
			\includegraphics[width=.33\linewidth]{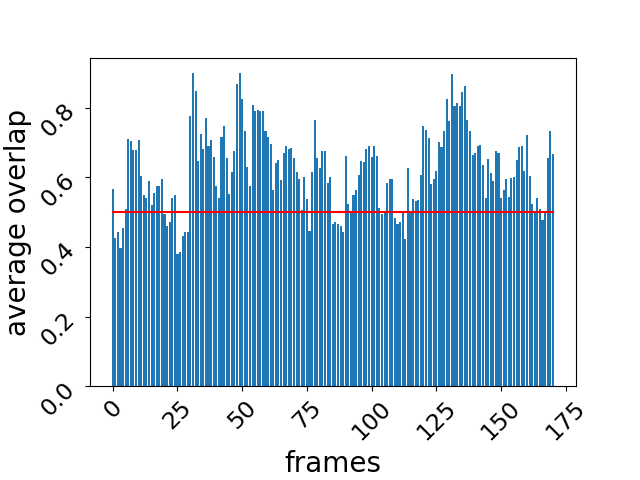}\hfill
			\includegraphics[width=.33\linewidth]{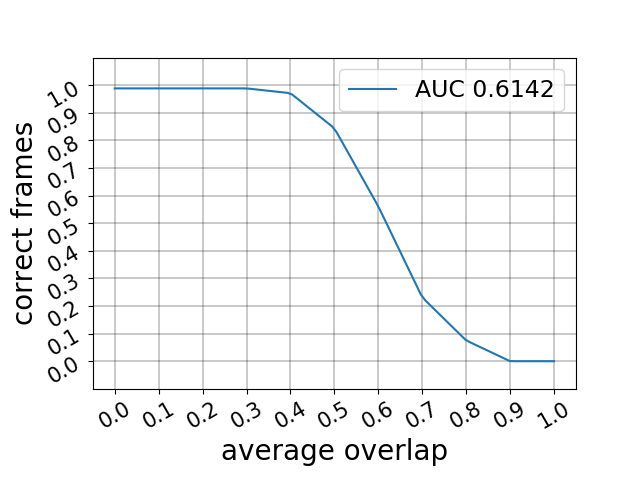}
			\includegraphics[width=.33\linewidth]{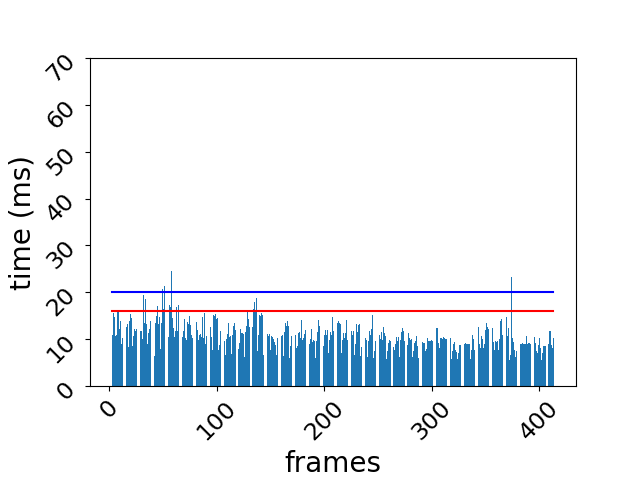}\hfill
			
		\end{subfigure}
		\caption*{Player 1}
		\begin{subfigure}{\linewidth}
			\includegraphics[width=.33\linewidth]{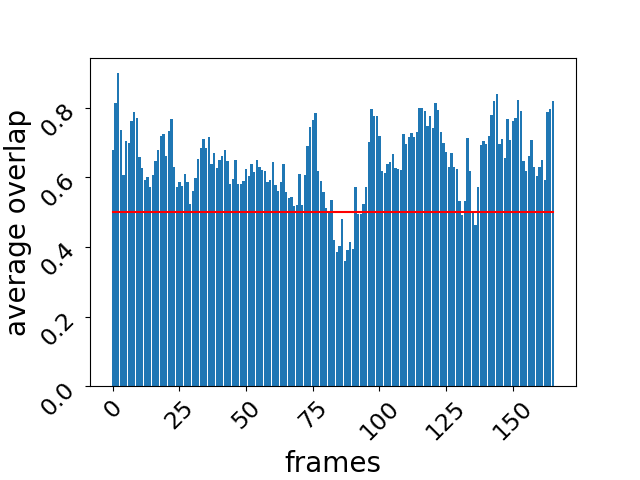}\hfill
			\includegraphics[width=.33\linewidth]{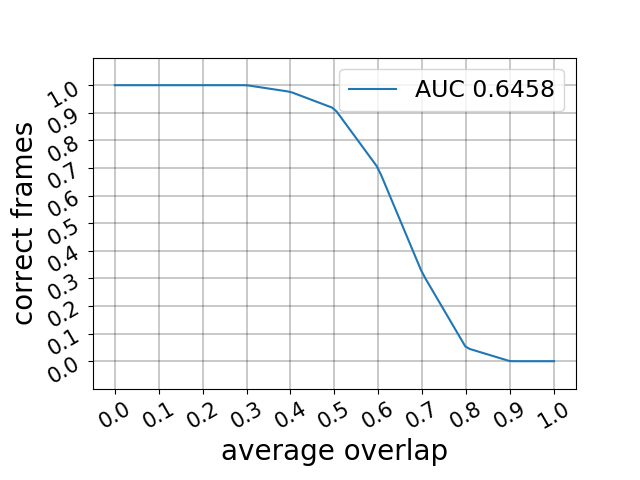}
			\includegraphics[width=.33\linewidth]{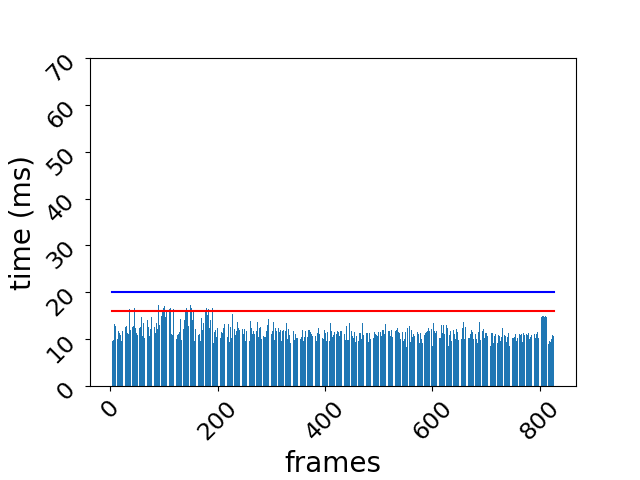}\hfill
			
		\end{subfigure}
		\caption*{Player 2}
		\begin{subfigure}{\linewidth}
			\includegraphics[width=.33\linewidth]{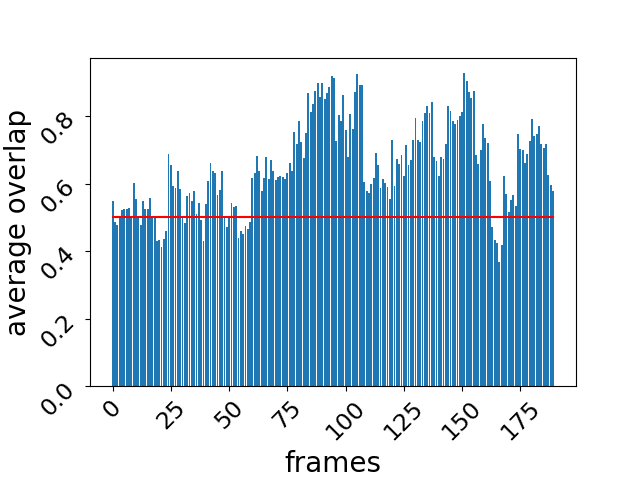}\hfill
			\includegraphics[width=.33\linewidth]{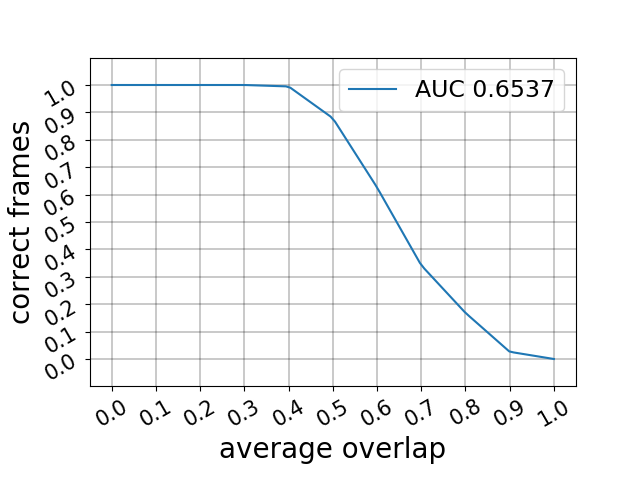}
			\includegraphics[width=.33\linewidth]{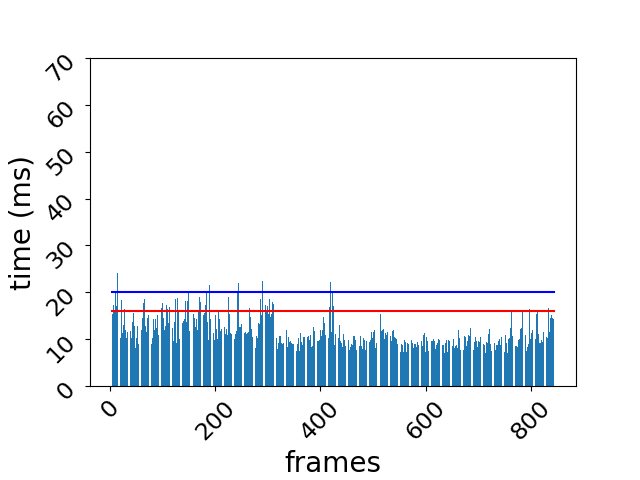}\hfill
			
		\end{subfigure}
		\caption*{Player 3}
		\begin{subfigure}{\linewidth}
			\includegraphics[width=.33\linewidth]{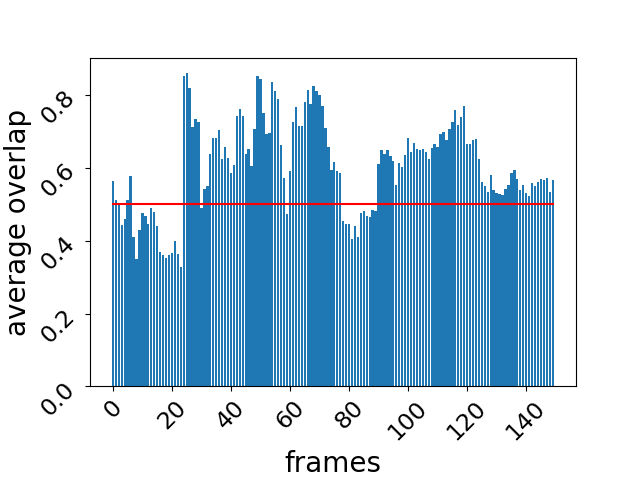}\hfill
			\includegraphics[width=.33\linewidth]{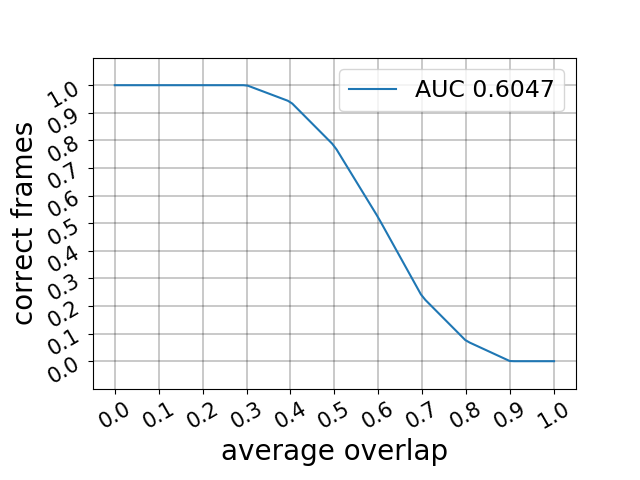}
			\includegraphics[width=.33\linewidth]{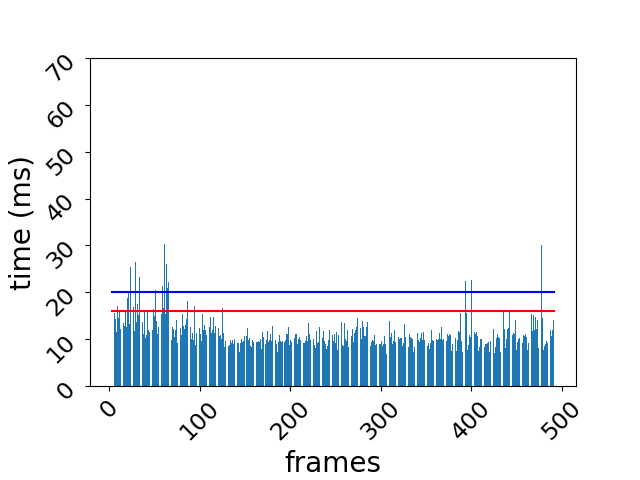}\hfill
		\end{subfigure}
		\caption*{Player 4}
		\caption{Average overlap, OPE curve and time per frame of each player using the hybrid tracker player}\label{fig:exp}
	\end{figure}
	
	Figure~\ref{fig:exp} contains the average overlap, OPE curve and execution time per frame figures for each one of the four players used in the evaluation.  
	The accuracy values are calculated against hand-labeled images, and we focused on segments of the video-sequence that we considered specially challenging.
	In the average overlap figures, a horizontal red line shows the threshold of 0.5 overlap, above which a detection is considered successful. As we can see, for all the players, the overlap is above that value most of the time, and when it is below, it does not drop too much below the threshold. Some of the areas of the videos below the threshold correspond to camera switches, or intervals in which the tracked player is not on camera. The OPE curve has a good shape, showing that the tracker achieves at least the minimum 0.5 overlap in between 80\% and 90\% of the frames. In addition, the summary value AUC is always around $0.6$, 
	which is always considered in the bibliography as an indicator of high precision (see Section~\ref{sec:ot}).
	These accuracy values are calculated against hand-labeled images, which allows making a more accurate evaluation than using videos labeled using a semiautomated labeling process with the aid of OpenCV. 
	In the following, the evaluation of the accuracy in this paper will be done against hand-labeled images. The downside of this approach is that hand-labeling takes significantly more time, thus, we had to focus the accuracy evaluation on segments of the video-sequence.
	
	Regarding the performance graphs, the red line corresponds to a frame rate of 60 fps, and the blue line corresponds to 50 fps. The results show that the frame time is almost always fast enough to be able to process at least 50 fps. However, the intervals of the video in which we cannot process more than 60 fps are rare. Notice that the tracking was done on a high definition video of a broadcasted video event with a frame rate of 60 fps. This input implied a higher computational cost, but, in exchange, it provided finer-grain details on every frame. The hardware used to execute our solution is the same workstation used for the tests described in Section~\ref{sec:motiv}.~\footnote{GTX 1080 Ti card, an Intel i7 8700k processor and 16 GB of RAM memory}


	\begin{table}[tb]
		\begin{tabular}{|l|l|l|l|l|}
			\hline
			& \textbf{Avg. Overlap}  & \textbf{Avg. Fps} & \textbf{Avg. AUC} & \textbf{Lost frames}\\ \hline
			\textbf{Player 1} & 0.620 & 91.75 & 0.610 & 2 \\ \hline
			\textbf{Player 2} & 0.653 & 84.98 & 0.651 & 0 \\ \hline
			\textbf{Player 3} & 0.650 & 86.65 & 0.660 & 0\\ \hline
			\textbf{Player 4} & 0.600 & 87.36 & 0.600 & 0\\ \hline
		\end{tabular}
		\caption{Average overlap, fps, AUC and Lost frames for the four tested players.}
		\label{tab:avgres}
	\end{table}
	
	Table~\ref{tab:avgres} shows the average accuracy, fps, AUC and lost frames for the video sequences of the four players. 
	The results show that our approach has an accuracy comparable to the state of the art techniques introduced in Section~\ref{sec:ot}. Notice that this is a remarkable achievement given the great difficulty of our case study, which is much higher than that of the benchmarks with which the state of the art approaches are usually evaluated. 
	The fps values show that the performance is really good, and the almost non-existent 
	lost frames prove that the tracker can deliver the tracking in real-time.
	
	\begin{figure}[tb]
		\begin{subfigure}{\linewidth}
			\includegraphics[width=.5\linewidth]{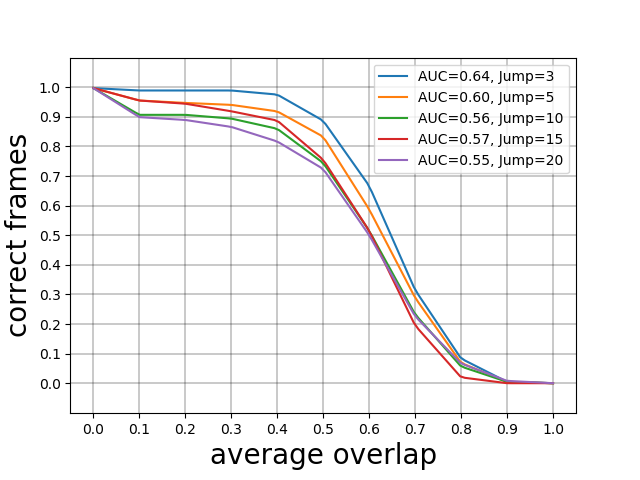}\hfill
			\includegraphics[width=.5\linewidth]{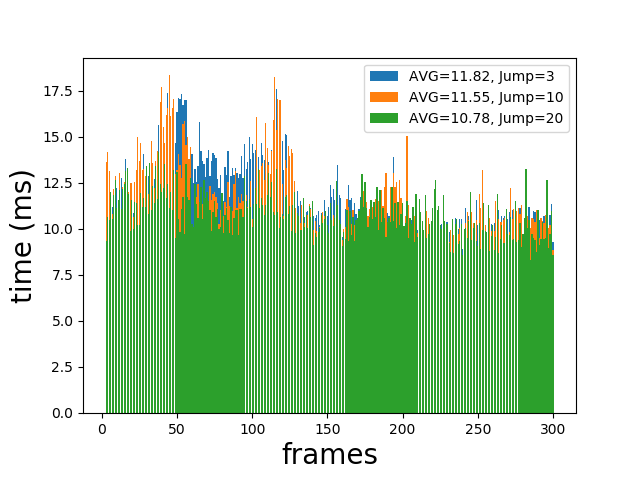}\hfill
		\end{subfigure}
		\caption{OPE curve and time per frame with different jump sizes}\label{fig:comp1}
		\begin{subfigure}{\linewidth}
			\includegraphics[width=.5\linewidth]{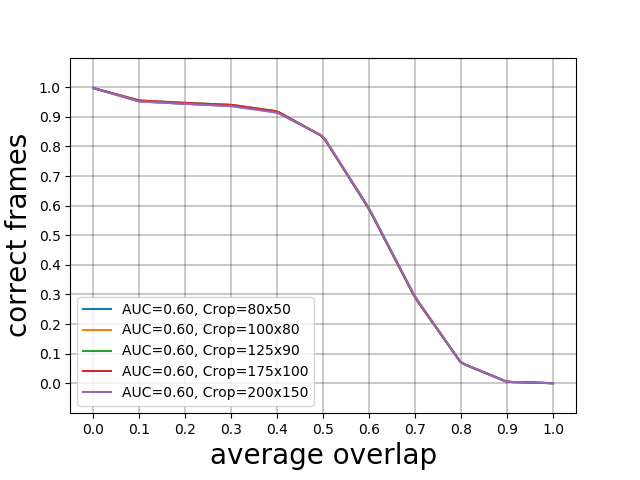}
			\includegraphics[width=.5\linewidth]{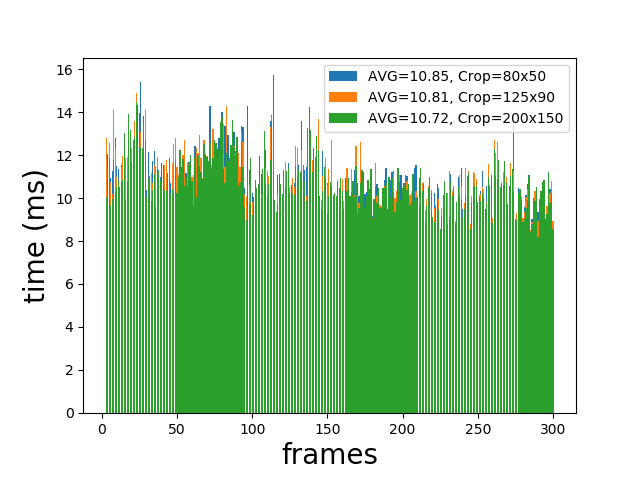}
		\end{subfigure}
		\caption{OPE curve and time per frame with different crop sizes}\label{fig:comp2}
		\caption*{}
	\end{figure}
	
	We finish this part of the evaluation discussing two parameters that may modify the behavior of our approach potentially impacting both the accuracy and the performance.
	
	The first one is the Faster-RCNN jump size between frames. 
	A small jump size increases the accuracy but it slows down the tracking, as it increases the frequency of execution of the Faster-RCNN model, which is accurate but slow. Figure \ref{fig:comp1} shows the OPE curve (left side) and the time per frame (right side) for different jump sizes. 
	A jump of three frames reaches the highest AUC, concretely $0.64$, but in exchange the time per frame needed is also considerably large and unstable. As the jump size increases, the area under the OPE curve is reduced as well as the time per frame, which also becomes more constant. 
	
	The second parameter of our approach that can be changed is the size of the crop of the frame made by the SSD model. Theoretically, a small crop size should favor a better accuracy, as the chances that there are two candidates for being the tracked player are smaller. Also, small crops of the frames will be processed faster, reducing the time per frame. The results shown in Figure~\ref{fig:comp2} for both the OPE curve and the time per frame indicate that the variation of this parameter does not have in practice a significant influence in none of both aspects.

	
	\subsection{Training time}~\label{sec:training} 
	
	Our method requires that the Faster-RCNN and the SSD networks are trained to detect an individual player. This training is done offline, but it requires a significant amount of computational resources.
	
	\begin{table}[h]
		\begin{tabular}{|l|l|l|l|l|l|}
			\hline
			& \textbf{Max. memory}  & \textbf{Execution Time} & \textbf{steps} & \textbf{loss} & \textbf{GPU} \\
			& \textbf{usage (Gb)}   &  \textbf{(hours)}     &       & &\\ \hline
			\textbf{Player 1} & 138.945 & 15 & 164166  & 0.0065 & nv\_k20\\ \hline
			\textbf{Player 2} & 139.800 & 20.4 & 200000 &  0.0014 & nv\_k20\\ \hline
			\textbf{Player 3} & 142.286 & 21.19 & 200000 &  0.0057 & nv\_k20\\ \hline
			\textbf{Player 4} & 144.545 & 20.6 & 200000 &  0.0033 & nv\_k20\\ \hline
		\end{tabular}
		\caption{Summary of training info for Faster-RCNN.}
		\label{tab:trainf}
	\end{table}
	
	\begin{table}[h]
		\begin{tabular}{|l|l|l|l|l|l|}
			\hline
			& \textbf{Max. memory}  & \textbf{Execution Time} & \textbf{steps} & \textbf{loss} & \textbf{GPU} \\
			& \textbf{usage (Gb)}   &  \textbf{(hours)}     &       & &\\ \hline
			\textbf{Player 1} & 143.163 & 46.9 & 200000 &  1.3549 & nv\_k20\\ \hline
			\textbf{Player 2} & 148.499 & 21.71 & 93768 &  1.3211 & nv\_k20\\ \hline
			\textbf{Player 3} & 151.271 & 22 & 95486 &  1.2148 & nv\_k20\\ \hline
			\textbf{Player 4} & 145.510 & 24 & 100152 & 1.0371 & nv\_k20\\ \hline
		\end{tabular}
		\caption{Summary of training info for SSD.}
		\label{tab:trains}
	\end{table}
	
	Tables~\ref{tab:trainf} and~\ref{tab:trains} summarize the training information for the two networks and the four players.
	Notice, that he training was done in a cluster node with a NVIDIA Tesla Kepler
	K20m 5 GB GDDR5, 2 x Intel Xeon E5-2660 Sandy Bridge-EP, and 64 GB of RAM memory.
	
	This training required to have a dataset of 1500 frames labeled with the location of the player, for each one of the four players. Doing this process by hand would have been infeasible. This is why we built a help tool that used KCF to automatically generate these datasets using a short sequence of videos where this technique worked well, and stopping the generation when the technique lost the tracked player. Although it was not a fully automatic process, it significantly reduced the time required to generate the dataset.
	
	
	
	
	

	\section{Conclusions}
	\label{sec:conc}
	
	The hybrid tracker presented in this paper combines two DL detection networks, Faster-RCNN and SSD, as well as a traditional math-based algorithm, like KCF, to produce a fast and accurate tracker for soccer players in broadcasted sport events. 
	
	The situation faced by the tracker is really complicated, as the video of the broadcasted sport event is not edited or preprocessed by us, and thus it contains total or partial occlusions of the tracked players, intervals of the sequence in which the tracked player is not on camera, and camera switches or movements. All these situations and others mentioned in the paper make this tracking really challenging.
	
	The experiments prove that our tracker can process 60 fps full HD videos at real-time while keeping a high degree of accuracy. This way, for instance, the achieved AUC is above 0.6 and the tracker can obtain an overlap with respect to hand-labeled frames of above 0.5 in more than 80\% of the frames.
	
	The proposed method is unsupervised, which requires that the DNNs are trained in advance to track a specific player. This allows the method not only to work without human supervision, but also to recover from camera switches and situations where the player has not been on camera for a few frames.
	
	As future work we are planing to group the detection of several players using the same Faster-RCNN network. This would decrease the time required to train the networks, and it would also enable the tracking of several players at the same time. Also, we plan to apply the same approach to other kinds of environments, such as tracking players in other sports or tracking persons of interest in video surveillance.
	
	\section*{Acknowledgements}
	This research was supported by the Ministerio de Econom\'ia, Industria y Competitividad  of  Spain  and  FEDER  funds  of  the  EU  (TIN2016-75845-P),  and by the Xunta de Galicia co-founded by the European Regional Development Fund (ERDF) under the Consolidation Programme of Competitive Reference Groups (ED431C 2017/04) as well as under the Centro Singular de Investigaci\'on de Galicia accreditation 2016-2019 (ED431G/01). We also acknowledge the Centro de Supercomputaci\'on de Galicia (CESGA) for the use of their computers.
	
	
	
	\bibliographystyle{elsarticle-harv} 
	\bibliography{references.bib}

\begin{thebibliography}{17}
\expandafter\ifx\csname natexlab\endcsname\relax\def\natexlab#1{#1}\fi
\providecommand{\url}[1]{\texttt{#1}}
\providecommand{\href}[2]{#2}
\providecommand{\path}[1]{#1}
\providecommand{\DOIprefix}{doi:}
\providecommand{\ArXivprefix}{arXiv:}
\providecommand{\URLprefix}{URL: }
\providecommand{\Pubmedprefix}{pmid:}
\providecommand{\doi}[1]{\href{http://dx.doi.org/#1}{\path{#1}}}
\providecommand{\Pubmed}[1]{\href{pmid:#1}{\path{#1}}}
\providecommand{\bibinfo}[2]{#2}
\ifx\xfnm\relax \def\xfnm[#1]{\unskip,\space#1}\fi
\bibitem[{Allen et~al.(2004)Allen, Xu and Jin}]{Allen:2004:OTU:1082121.1082122}
\bibinfo{author}{Allen, J.G.}, \bibinfo{author}{Xu, R.Y.D.},
  \bibinfo{author}{Jin, J.S.}, \bibinfo{year}{2004}.
\newblock \bibinfo{title}{Object tracking using camshift algorithm and multiple
  quantized feature spaces}, in: \bibinfo{booktitle}{Proceedings of the
  Pan-Sydney Area Workshop on Visual Information Processing},
  \bibinfo{publisher}{Australian Computer Society, Inc.},
  \bibinfo{address}{Darlinghurst, Australia, Australia}. pp.
  \bibinfo{pages}{3--7}.
\newblock \URLprefix \url{http://dl.acm.org/citation.cfm?id=1082121.1082122}.
\bibitem[{{Babenko} et~al.(2009){Babenko}, {Yang} and {Belongie}}]{mil}
\bibinfo{author}{{Babenko}, B.}, \bibinfo{author}{{Yang}, M.},
  \bibinfo{author}{{Belongie}, S.}, \bibinfo{year}{2009}.
\newblock \bibinfo{title}{Visual tracking with online multiple instance
  learning}, in: \bibinfo{booktitle}{2009 IEEE Conference on Computer Vision
  and Pattern Recognition}, pp. \bibinfo{pages}{983--990}.
\bibitem[{Bertinetto et~al.(2016)Bertinetto, Valmadre, Henriques, Vedaldi and
  Torr}]{bertinetto2016fully}
\bibinfo{author}{Bertinetto, L.}, \bibinfo{author}{Valmadre, J.},
  \bibinfo{author}{Henriques, J.F.}, \bibinfo{author}{Vedaldi, A.},
  \bibinfo{author}{Torr, P.H.}, \bibinfo{year}{2016}.
\newblock \bibinfo{title}{Fully-convolutional siamese networks for object
  tracking}, in: \bibinfo{booktitle}{European conference on computer vision},
  \bibinfo{organization}{Springer}. pp. \bibinfo{pages}{850--865}.
\bibitem[{Comaniciu and Meer(2002)}]{Comaniciu:2002:MSR:513073.513076}
\bibinfo{author}{Comaniciu, D.}, \bibinfo{author}{Meer, P.},
  \bibinfo{year}{2002}.
\newblock \bibinfo{title}{Mean shift: A robust approach toward feature space
  analysis}.
\newblock \bibinfo{journal}{IEEE Trans. Pattern Anal. Mach. Intell.}
  \bibinfo{volume}{24}, \bibinfo{pages}{603--619}.
\newblock \URLprefix \url{https://doi.org/10.1109/34.1000236},
  \DOIprefix\doi{10.1109/34.1000236}.
\bibitem[{{G.B.. A. Karhler}(2008)}]{opencv}
\bibinfo{author}{{G.B.. A. Karhler}}, \bibinfo{year}{2008}.
\newblock \bibinfo{title}{Learning opencv, computer vision with the opencv
  library}.
\bibitem[{Giancola et~al.(2018)Giancola, Amine, Dghaily and
  Ghanem}]{giancola2018soccernet}
\bibinfo{author}{Giancola, S.}, \bibinfo{author}{Amine, M.},
  \bibinfo{author}{Dghaily, T.}, \bibinfo{author}{Ghanem, B.},
  \bibinfo{year}{2018}.
\newblock \bibinfo{title}{Soccernet: A scalable dataset for action spotting in
  soccer videos}.
\newblock \bibinfo{journal}{arXiv preprint arXiv:1804.04527} .
\bibitem[{Guo et~al.(2017)Guo, Feng, Zhou, Huang, Wan and
  Wang}]{guo2017learning}
\bibinfo{author}{Guo, Q.}, \bibinfo{author}{Feng, W.}, \bibinfo{author}{Zhou,
  C.}, \bibinfo{author}{Huang, R.}, \bibinfo{author}{Wan, L.},
  \bibinfo{author}{Wang, S.}, \bibinfo{year}{2017}.
\newblock \bibinfo{title}{Learning dynamic siamese network for visual object
  tracking}, in: \bibinfo{booktitle}{Proceedings of the IEEE International
  Conference on Computer Vision}, pp. \bibinfo{pages}{1763--1771}.
\bibitem[{He et~al.(2018)He, Luo, Tian and Zeng}]{he2018twofold}
\bibinfo{author}{He, A.}, \bibinfo{author}{Luo, C.}, \bibinfo{author}{Tian,
  X.}, \bibinfo{author}{Zeng, W.}, \bibinfo{year}{2018}.
\newblock \bibinfo{title}{A twofold siamese network for real-time object
  tracking}, in: \bibinfo{booktitle}{Proceedings of the IEEE Conference on
  Computer Vision and Pattern Recognition}, pp. \bibinfo{pages}{4834--4843}.
\bibitem[{Henriques et~al.(2014)Henriques, Caseiro, Martins and Batista}]{kcf}
\bibinfo{author}{Henriques, J.}, \bibinfo{author}{Caseiro, R.},
  \bibinfo{author}{Martins, P.}, \bibinfo{author}{Batista, J.},
  \bibinfo{year}{2014}.
\newblock \bibinfo{title}{High-speed tracking with kernelized correlation
  filters}.
\newblock \bibinfo{journal}{IEEE Transactions on Pattern Analysis and Machine
  Intelligence} \bibinfo{volume}{37}.
\newblock \DOIprefix\doi{10.1109/TPAMI.2014.2345390}.
\bibitem[{Kristan et~al.(2016)Kristan, Matas, Leonardis, Vojir, Pflugfelder,
  Fernandez, Nebehay, Porikli and \v{C}ehovin}]{VOT_TPAMI}
\bibinfo{author}{Kristan, M.}, \bibinfo{author}{Matas, J.},
  \bibinfo{author}{Leonardis, A.}, \bibinfo{author}{Vojir, T.},
  \bibinfo{author}{Pflugfelder, R.}, \bibinfo{author}{Fernandez, G.},
  \bibinfo{author}{Nebehay, G.}, \bibinfo{author}{Porikli, F.},
  \bibinfo{author}{\v{C}ehovin, L.}, \bibinfo{year}{2016}.
\newblock \bibinfo{title}{A novel performance evaluation methodology for
  single-target trackers}.
\newblock \bibinfo{journal}{IEEE Transactions on Pattern Analysis and Machine
  Intelligence} \bibinfo{volume}{38}, \bibinfo{pages}{2137--2155}.
\newblock \DOIprefix\doi{10.1109/TPAMI.2016.2516982}.
\bibitem[{Li et~al.(2018)Li, Yan, Wu, Zhu and Hu}]{li2018high}
\bibinfo{author}{Li, B.}, \bibinfo{author}{Yan, J.}, \bibinfo{author}{Wu, W.},
  \bibinfo{author}{Zhu, Z.}, \bibinfo{author}{Hu, X.}, \bibinfo{year}{2018}.
\newblock \bibinfo{title}{High performance visual tracking with siamese region
  proposal network}, in: \bibinfo{booktitle}{Proceedings of the IEEE Conference
  on Computer Vision and Pattern Recognition}, pp. \bibinfo{pages}{8971--8980}.
\bibitem[{Li and Zhang(2019)}]{cite5}
\bibinfo{author}{Li, Y.}, \bibinfo{author}{Zhang, X.}, \bibinfo{year}{2019}.
\newblock \bibinfo{title}{Siamvgg: Visual tracking using deeper siamese
  networks}.
\newblock \bibinfo{journal}{arXiv preprint arXiv:1902.02804} .
\bibitem[{Manafifard et~al.(2017)Manafifard, Ebadi and Moghaddam}]{cite11}
\bibinfo{author}{Manafifard, M.}, \bibinfo{author}{Ebadi, H.},
  \bibinfo{author}{Moghaddam, H.A.}, \bibinfo{year}{2017}.
\newblock \bibinfo{title}{A survey on player tracking in soccer videos}.
\newblock \bibinfo{journal}{Computer Vision and Image Understanding}
  \bibinfo{volume}{159}, \bibinfo{pages}{19--46}.
\bibitem[{Svensson(2010)}]{svensson2010target}
\bibinfo{author}{Svensson, D.}, \bibinfo{year}{2010}.
\newblock \bibinfo{title}{Target tracking in complex scenarios}.
\newblock \bibinfo{publisher}{Chalmers University of Technology}.
\bibitem[{Valmadre et~al.(2017)Valmadre, Bertinetto, Henriques, Vedaldi and
  Torr}]{valmadre2017end}
\bibinfo{author}{Valmadre, J.}, \bibinfo{author}{Bertinetto, L.},
  \bibinfo{author}{Henriques, J.}, \bibinfo{author}{Vedaldi, A.},
  \bibinfo{author}{Torr, P.H.}, \bibinfo{year}{2017}.
\newblock \bibinfo{title}{End-to-end representation learning for correlation
  filter based tracking}, in: \bibinfo{booktitle}{Proceedings of the IEEE
  Conference on Computer Vision and Pattern Recognition}, pp.
  \bibinfo{pages}{2805--2813}.
\bibitem[{Wang et~al.(2018)Wang, Teng, Xing, Gao, Hu and
  Maybank}]{wang2018learning}
\bibinfo{author}{Wang, Q.}, \bibinfo{author}{Teng, Z.}, \bibinfo{author}{Xing,
  J.}, \bibinfo{author}{Gao, J.}, \bibinfo{author}{Hu, W.},
  \bibinfo{author}{Maybank, S.}, \bibinfo{year}{2018}.
\newblock \bibinfo{title}{Learning attentions: residual attentional siamese
  network for high performance online visual tracking}, in:
  \bibinfo{booktitle}{Proceedings of the IEEE Conference on Computer Vision and
  Pattern Recognition}, pp. \bibinfo{pages}{4854--4863}.
\bibitem[{Wu et~al.(2013)Wu, Lim and Yang}]{WuLimYang13}
\bibinfo{author}{Wu, Y.}, \bibinfo{author}{Lim, J.}, \bibinfo{author}{Yang,
  M.H.}, \bibinfo{year}{2013}.
\newblock \bibinfo{title}{Online object tracking: A benchmark}, in:
  \bibinfo{booktitle}{IEEE Conference on Computer Vision and Pattern
  Recognition (CVPR)}, pp. \bibinfo{pages}{2411--2418}.

\end{thebibliography}
	
	
	
	
	
\end{document}